\documentclass[letterpaper, 10 pt, conference]{ieeeconf}
\IEEEoverridecommandlockouts
\usepackage{graphicx}
\def\BibTeX{{\rm B\kern-.05em{\sc i\kern-.025em b}\kern-.08em
    T\kern-.1667em\lower.7ex\hbox{E}\kern-.125emX}}
\usepackage[font=small,labelfont=bf,tableposition=top]{caption}
\usepackage[utf8]{inputenc}
\usepackage[T1]{fontenc}
\usepackage{multirow}
\usepackage{graphicx}
\usepackage{gensymb}
\usepackage{cite}
\usepackage{amsmath} 
\usepackage{colortbl}
\usepackage{tikz}
\usepackage{pgfplots}
\usepackage{subfig}
\usepackage{amssymb}
\usepackage{CJKutf8}
\usepackage{algorithm}
\usepackage{algpseudocode}
\usepackage{booktabs}

\makeatletter
\let\NAT@parse\undefined
\makeatother
\usepackage{xcolor}
\definecolor{rblue}{rgb}{0,0.5,1}
\definecolor{hollywoodcerise}{rgb}{0.96, 0.0, 0.63}
\definecolor{lasallegreen}{rgb}{0.03, 0.47, 0.19}
\definecolor{hanpurple}{rgb}{0.32, 0.09, 0.98}
\definecolor{green(pigment)}{rgb}{0.0, 0.65, 0.31}
\usepackage[pagebackref=false,breaklinks=true,colorlinks,bookmarks=false]{hyperref}
\hypersetup{colorlinks,linkcolor={red},citecolor={hanpurple},urlcolor={magenta}}  

\definecolor{mygray}{gray}{.9}
\definecolor{mygreen}{RGB}{93,174,86}

\usepackage{bbding}

\begin{document}
\title{\LARGE \bf
L2G-Map: Local-to-Global Mapping via Hierarchical Diffusion Refinement and Elliptical Bayesian Fusion}
\author{Siyu Li$^{1,2,*}$, Xinying Hong$^{3,*}$, Fei Teng$^{2}$, Kang Zeng$^{3}$, Hao Shi$^{4}$, Beiping Hou$^{1}$,\\Zhiyong Li$^{2,3,\dag}$, and Kailun Yang$^{2,\dag}$%
\thanks{This work was supported in part by the National Natural Science Foundation of China (No. 62473139), in part by the Hunan Provincial Research and Development Project (Grant No. 2025QK3019), in part by the State Key Laboratory of Autonomous Intelligent Unmanned Systems (the opening project number ZZKF2025-2-10), by National Natural Science Foundation of China (No.U23A20341), and Major Scientific and Technological Research Plan of Hunan Province (No.2025QK1005).}
\thanks{$^{1}$S. Li and B. Hou are with the School of Automation and Electrical Engineering, Zhejiang University of Science and Technology, China.}
\thanks{$^{2}$S. Li, Z. Li, and K. Yang are with the School of Artificial Intelligence and Robotics and the National Engineering Research Center of Robot Visual Perception and Control Technology, Hunan University, China.}
\thanks{$^{3}$X. Hong, K. Zeng, and Z. Li are with the College of Computer Science and Electronic Engineering, Hunan University, China.}
\thanks{$^{4}$S. Hao is with Ant Group, China.}%
\thanks{$^{*}$Equal contribution.}
\thanks{{\dag}Corresponding authors (e-mail: firstname.lastname@hnu.edu.cn).}
}

\maketitle

\begin{abstract}
Offline high-definition maps provide essential geometric and topological priors for autonomous driving systems. Pure-vision solutions have become the predominant paradigm for offline mapping due to their cost-effectiveness and scalability. However, local-to-global mapping under visual conditions confronts two fundamental challenges: single-shot local observations are susceptible to viewpoint variation and environmental interference, leading to geometric deviations, while multi-source local information exhibits heterogeneous confidence, rendering globally consistent aggregation difficult. To address these, this paper proposes L2G-Map, a framework comprising hierarchical prior diffusion refinement and elliptical space Bayesian fusion. The former jointly embeds temporal context and centerline priors to guide structure completion and topology recovery during denoising, alleviating the information incompleteness inherent in pure-vision settings. The latter incorporates an adaptive weighting strategy driven by elliptical distance propagation, enabling probabilistically optimal aggregation of multi-source information under the Bayesian posterior update paradigm. Extensive experiments on nuScenes and Argoverse benchmark datasets verify the effectiveness of L2G‑Map. The proposed refinement component yields consistent local map accuracy improvements across different datasets. Under sensor-degraded conditions, a $3.27\%$ mIoU gain is achieved. Furthermore, the adaptive fusion component significantly enhances the accuracy of global maps. The fused global map can be flexibly embedded into different online map models, yielding an $18.26\%$ mIoU improvement in semantic map construction and a $20.00\%$ enhancement in vectorized map construction, demonstrating the overall advantages of the proposed closed‑loop pipeline. Source code will be available at \url{https://github.com/lynn-yu/L2G-Map}.

\end{abstract}

\IEEEpeerreviewmaketitle

\section{Introduction}
Offline High-Definition (HD) maps serve as a fundamental component in autonomous driving architectures, offering strong priors that facilitate online map perception while simultaneously delivering long-range contextual information to support downstream tasks including trajectory forecasting and behavior prediction~\cite{mappath}. 
Owing to the low cost of visual sensors, offline HD maps constructed from visual images have emerged as the predominant paradigm, which generally involves local map generation and global map fusion~\cite{yuan2026unimapgen}. 

\begin{figure}[tb]
    \centering
    \includegraphics[scale=0.50]{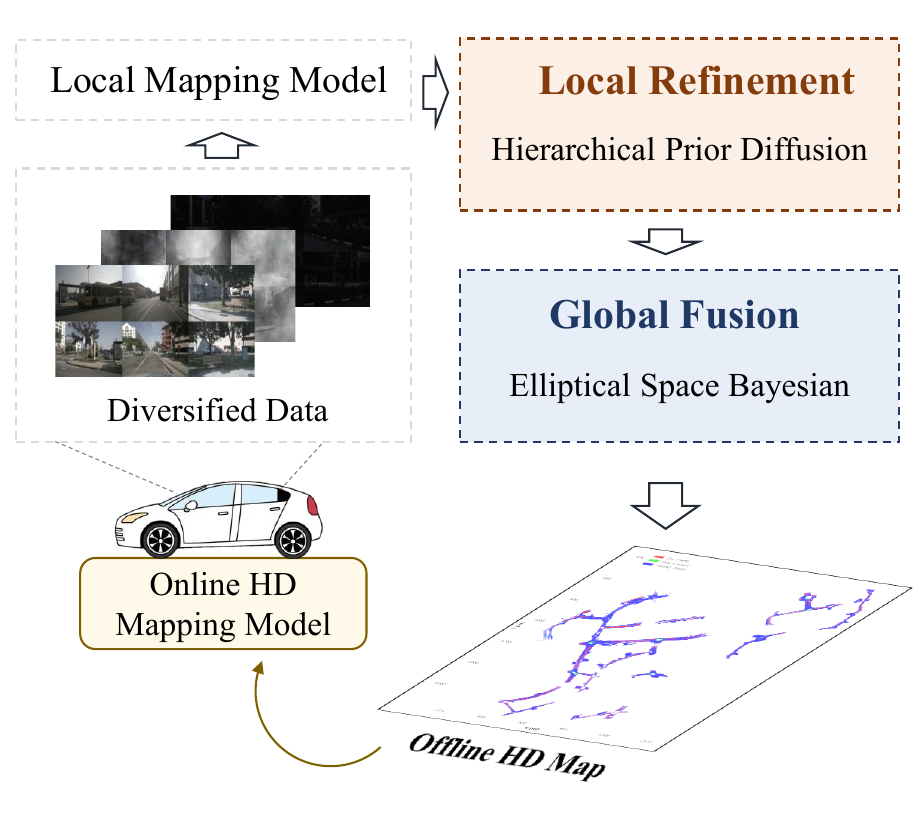}
    \vskip-3ex
    \caption{Overview of L2G-Map. Through hierarchical prior diffusion refinement and elliptical space Bayesian fusion, the proposed framework constructs offline HD maps from vehicle sensory data, which in turn provide feedback to online map models to continuously strengthen perception capabilities.
    }
    \label{fig:intro}
    \vskip-4ex
\end{figure}
For local map generation, online mapping models represent a promising choice~\cite{philion2020lift,li2022hdmapnet,jiang2024p,li2024genmapping}.
These models can directly convert images captured by onboard vision sensors into local maps. Although they demonstrate strong capabilities in real-time mapping, their performance degrades in diverse real-world complex scenarios, such as adverse weather conditions and sensor degradation.
Given that the quality of local maps fundamentally dictates the performance of global map construction, ensuring locally accurate map generation under diverse conditions is of paramount importance.
Furthermore, during the process of fusing local maps into a global map, ghosting artifacts may arise due to model errors and noise. 
An ambiguous global map fails to provide accurate prior information for online map perception and downstream tasks.
Therefore, designing an efficient fusion method also remains a key challenge in offline map construction.

To address these challenges, this paper proposes a global map construction method that proceeds from local refinement to global fusion, as shown in Fig.~\ref{fig:intro}.
Specifically, a Hierarchical Prior Diffusion Refinement (HPDR) component is proposed to realize local refinement. 
Given the coarse local maps generated by online mapping models, a diffusion-based framework is introduced to model the underlying implicit distribution of map features. 
By leveraging temporal cues and centerline map priors to steer the denoising procedure, the approach gradually recovers optimal maps from the learned noise distribution.
Meanwhile, an Elliptical Space Bayesian Fusion (ESBF) component is designed to construct global maps. 
Inspired by human driving reasoning, a confidence estimation strategy based on elliptical distance propagation is developed to model the uncertainty arising from local map generation. 
This strategy is incorporated into the Bayesian posterior update procedure, facilitating efficient aggregation of multi-source local maps.

Experimental results on both local mapping and global fusion tasks demonstrate the effectiveness of the proposed method. 
The proposed local refinement module demonstrates consistent and significant accuracy gains across a variety of online map backbones, yielding a $3.27\%$ improvement under sensor‑degraded scenarios. 
The global fusion component, in turn, achieves efficient and reliable aggregation of local map inputs.
Moreover, the resulting global map can be seamlessly integrated into both semantic and vector online mapping models, substantially enhancing their perceptual performance.
Our contributions are summarized as follows:
\begin{enumerate}
\item We introduce L2G-Map, a unified local-to-global framework that formulates offline HD map construction as an integrated pipeline.
\item We propose a hierarchical prior diffusion refinement component leveraging different priors to recover accurate local maps, and an elliptical space Bayesian fusion component with distance-propagation confidence weighting for optimal global map aggregation.
\item The proposed integrated map construction architecture is capable of building global maps under diverse environmental conditions, while closing the loop to empower online HD map models.
\end{enumerate}

\section{Related Work}
\noindent\textbf{Online HD Mapping.}
Online HD maps are typically constructed from a Bird's Eye View (BEV). A core technical challenge lies in achieving view transformation from multi-view monocular camera data to the BEV space.
LSS~\cite{philion2020lift} establishes explicit geometric correspondence via depth estimation.
BEVFormer~\cite{li2024bevformer} leverages temporal attention for feature aggregation. 
HDMapnet~\cite{li2022hdmapnet} and P-MapNet~\cite{jiang2024p} adopt a multi-layer perceptron to implicitly learn this transformation process.
GenMapping~\cite{li2024genmapping} employs inverse perspective mapping to enhance the generalization ability.
Although these models achieve good accuracy in real-time mapping, they face significant challenges when constructing local maps from diverse crowdsourced data.
To achieve more stable offline local map learning, this paper aims to employ diverse prior information to enhance map accuracy.

\begin{figure*}[tb]
    \centering
    \includegraphics[width=1.0\textwidth]{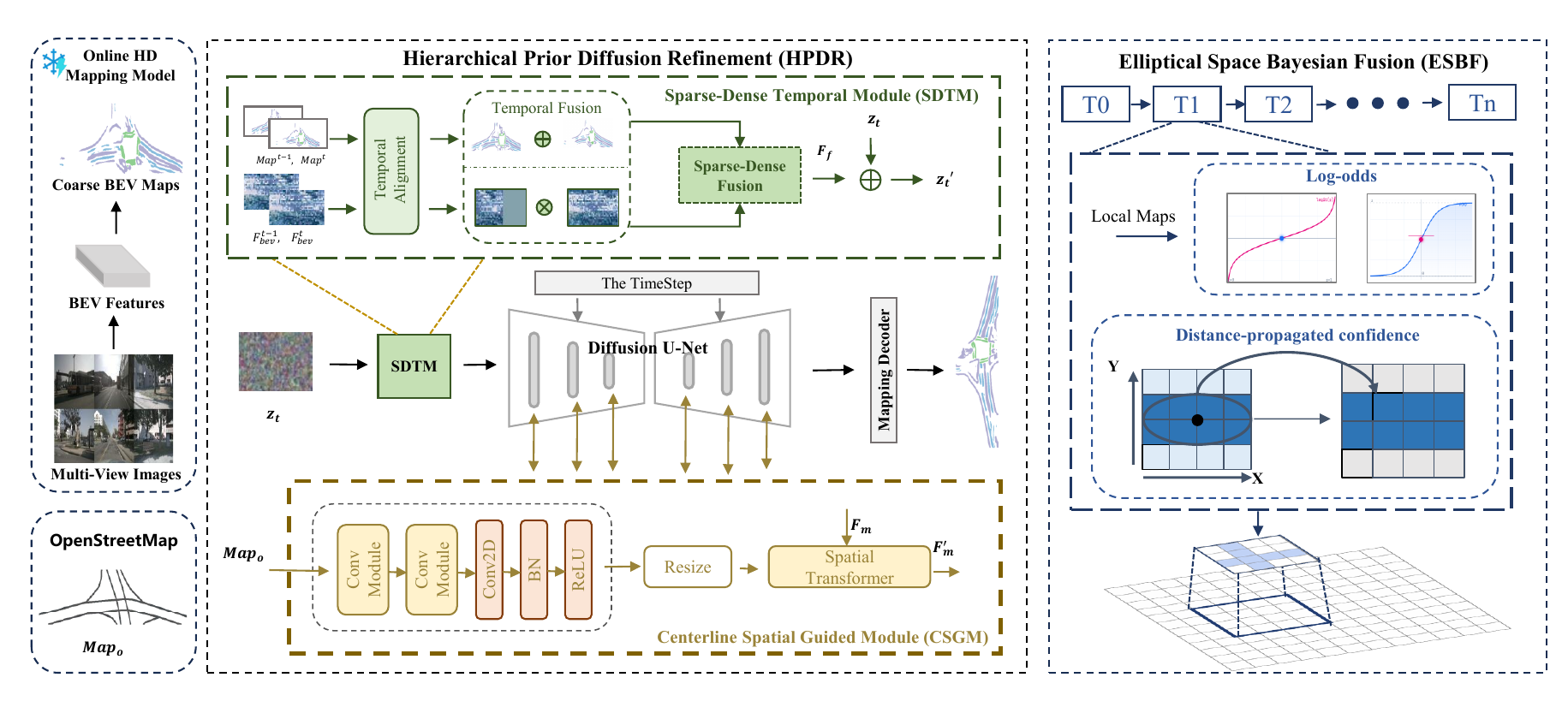}
    \vskip-2ex
    \caption{Overview of the proposed L2G framework.The onboard multi-view images are first processed by the local mapping module to produce coarse maps, which are then refined by the local refinement module and finally fused by the global fusion module to generate the offline global map.}
    \label{fig:structure_of_TS-CGNet}
    \vskip-4ex
\end{figure*}

\noindent\textbf{Prior-Guided Map Construction.}
Prior information, such as Standard Definition Maps (SDMap) from OpenStreetMap (OSM)~\cite{haklay2008openstreetmap}, significantly enhances mapping performance. 
Methods like P-MapNet~\cite{jiang2024p} and BLOS-BEV~\cite{wu2024blos} incorporate weakly aligned SDMap priors to guide online generators and extend perception ranges. 
Simultaneously, historical traversals provide temporal priors to mitigate sensor sparsity. 
NMP~\cite{xiong2023neural} and GlobalMapNet~\cite{shi2024globalmapnet} construct global priors from previous passes, whereas StreamMapNet~\cite{yuan2024streammapnet} and DTCLMapper~\cite{li2024dtclmapper} leverage streaming strategies and consistency learning to preserve spatiotemporal information.
AMap~\cite{li2026amap} distills future temporal frames into a current-frame student model to enable look-ahead perception at zero inference-time cost.
Despite the proven benefits of both map and temporal priors, research exploring their combined potential remains scarce. 

\noindent\textbf{Condition Diffusion Models.}
Generative models, such as VAE~\cite{kingma2013auto,vahdat2020nvae} and Masked Auto-Encoder (MAE)~\cite{he2022masked,liu2205mixed}, have been widely applied to image reconstruction and offboard map learning~\cite{jiang2024p, zhu2023mapprior}. 
For instance, P-MapNet~\cite{jiang2024p} employs an MAE to infer unobserved regions, though its adaptability remains limited in complex scenarios. 
Recently, diffusion models~\cite{dhariwal2021diffusion,ho2020denoising,rombach2022high} and conditional frameworks like ControlNet~\cite{zhang2023adding} and DDP~\cite{ji2023ddp} have demonstrated superior performance in generating high-quality images and semantic masks by denoising in latent spaces.
DiffMap~\cite{jia2024diffmap} only uses semantic features as conditional guidance for the diffusion model to construct semantic maps, overlooking the effectiveness of multi-source prior information, which is a direction worth exploring.
This paper explores a novel framework that integrates dual prior knowledge.

\noindent\textbf{Global Map Construction.}
Global map construction generally involves optimizing the fusion of local maps through alignment based on estimated vehicle poses.
BEV-SLAM~\cite{ross2022bev} utilized BEV semantic matching to optimize visual SLAM poses, and directly accumulates local maps for fusion.
UniMapGen~\cite{yuan2026unimapgen} adopted a joint mapping framework that integrates satellite and visual imagery for map construction.
CSMapping~\cite{qiao2025csmapping} employed latent diffusion priors for crowdsourced semantic mapping from noisy observations, focusing on map completion via latent-space optimization. Different from CSMapping, this paper applies diffusion models for local map refinement conditioned on hierarchical priors, rather than global completion. 
Additionally, our framework establishes a closed loop from offline global maps back to online perception.


\section{Methodology}

\subsection{Overview}
As shown in Fig.~\ref{fig:intro}, L2G-Map is designed to first generate local maps from sequential road data collected by the vehicle, and then fuse them to construct a high-quality global map. 
Among them, to generate complete local maps from data collected in diverse environments, a Hierarchical Prior Diffusion Refinement (HPDR) module is proposed, which synergistically leverages multiple priors to iteratively generate the optimal local map through noise diffusion. Then, considering the uncertainty of consecutive local maps, an Elliptical Space Bayesian Fusion (ESBF) module is designed, which introduces elliptical distance propagated confidence embedded into Bayesian updates for adaptive fusion.

\subsection{Local Mapping: Hierarchical Prior Diffusion Refinement}
Considering the increasing maturity of online mapping models~\cite{li2022hdmapnet,jiang2024p}, this paper adopts such a model to generate coarse local maps from the collected multi-view images, which includes image encoding, view transformer, and decoder modules. 
Image encoding takes perspective images $I_n$ as input to learn semantic features and spatial relationships. 
The view transformer module jointly learns semantic and geometric spatial relationships to convert visual features into BEV representations $F_{bev}$.
Finally, the decoder module can generate coarse local maps $Map$.

However, when confronted with complex environmental data such as adverse weather conditions, local maps generated by online mapping models are prone to artifacts such as fractures, resulting in incomplete map reconstruction.
Given that the quality of local maps serves as the foundation for the entire high-definition map, failure to mitigate this issue would allow errors to accumulate progressively, thereby exerting a substantial adverse impact on the quality of the resulting global map.
To this end, with the aim of enhancing the quality of coarse local maps, this section proposes a sparse-dense temporal module and a centerline spatial guided module, which progressively learn complete map structures through the incorporation of multi-level prior knowledge within the conditional diffusion learning process.

\textbf{1) Effective Conditional Diffusion Architecture:} 
To enhance learning efficiency, the diffusion employs a lightweight U-Net.
$f_\theta$ consists of $27$ modules, with $8$ of them dedicated to upsampling and downsampling operations to adjust the feature map sizes. 
Among the remaining processing modules, the $15$ modules are composed of ResBlocks~\cite{rombach2022high} and Spatial Transformers~\cite{jaderberg2015spatial}, which are concatenated via skip connections. 
The Spatial Transformers learn the relationship between the noise latent $z_t$ and each timestep $t$.
The central $4$ modules retain the original ResBlock design to ensure stability and consistency in the core region. 

The dual conditioning modules, incorporating temporal-spatial priors and centerline priors respectively, are sequentially applied in the diffusion learning process to collaboratively guide local map re-learning. 

\textbf{2) Sparse-Dense Temporal Module:} 
Given the inherent sparsity and uncertainty of single-frame maps, this section leverages dense BEV features and temporal information as complementary priors to facilitate re-learning, capitalizing on their rich spatial and temporal context.

Specifically, we first separately fuse relevant information from historical frames for both sparse maps and dense features.
The historical frame map $\text{Map}^{t-1}$ can be translated to the current frame by $\text{Map}^{t-1'}=\text{Tr}(\text{Map}^{t-1})$.
Subsequently, they are fused with the current map $\text{Map}^{t}$.
\vspace{-0.5em}
\begin{equation}
   F_m = \text{ReLU} \left( \varphi_1 (\text{Map}^{t-1'} + \text{Map}^{t}) \right),
   \vspace{-0.6em}
\end{equation}
where $F_m$ indicates the map features that are aligned for feature fusion and $\varphi_1$ denotes the $3{\times3}$ convolution. \text{ReLU} denotes the rectified linear unit activation.

Subsequently, the process involves the temporal fusion of BEV features.
Aligned temporal features, being static, exhibit similar spatial structures. Direct summation may introduce redundancy and mislead learning. Hence, concatenation with channel-wise adaptive learning is adopted to fuse them, effectively compensating for missing information while avoiding redundancy.
\vspace{-0.5em}
\begin{equation}
     F_e = \varphi_2(\text{Concat}( (\text{Tr}(\text{F}_{bev}^{t-1}),\text{F}_{bev}^{t}))),
     \vspace{-0.2em}
\end{equation}
where $F_be$ indicates the fused feature, $\text{F}_{b}^{t-1}$ and $\text{F}_{b}^{t}$ represent features from online mapping models. $\text{Tr}$ is is an alignment operation.
$\varphi_2$ is composed of $1\times1$ convolution and ReLu activation functions.

The features obtained from two levels are fused into a complete feature $F_{f}$ through the convolution block $\varphi_3$, obtaining a rich characterization of semantic mapping:
\vspace{-0.3em}
\begin{equation}
    F_f = \varphi_3(F_m+F_e).
\vspace{-0.4em}   
\end{equation}
Finally, this prior feature $F_f$ is integrated into the latent noise distribution, implicitly guiding the diffusion denoising process.
\begin{equation}
    z'_t = z_t + F_f.
\end{equation}

\textbf{3) Centerline Spatial Guided Module:}
Centerline maps from OpenStreetMap (OSM)~\cite{haklay2008openstreetmap} are highly reliable and robust, remaining unaffected by weather or environmental conditions.
Consequently, this section employs centerline maps as control conditions for the denoising process, progressively guiding the process of re-learning.

Since OSM maps are binary representations, a convolutional network is employed to extract local features.
This network includes $4\times4$ dilated convolutions (DCV) followed by activation functions (ReLU). Then, deep detailed features $F_o$ can be obtained through $2\times2$ dilated convolutions followed by BatchNorm2D (BN) and activation functions (ReLU).
\vspace{-0.4em}
\begin{equation}
    F_o = \text{ReLU}(\text{BN}(\text{DCV}([\text{ReLU}(\text{DCV}(Map_o))]_{\times2}))).
    \vspace{-0.4em}
\end{equation}
The centerline features $F_o$ are then embedded into the diffusion U-Net architecture, specifically injected at every hierarchical level.
To account for positional deviations in centerline maps, the cross-attention mechanism in the $\text{SpatialTrans}()$ module is employed between the denoising features and centerline features for controlled learning: 
\vspace{-0.3em}
\begin{equation}
    F'_m = \text{SpatialTrans}(F_o,F_m)+F_m.
    \vspace{-0.3em}
\end{equation}
The enhanced denoising features $F'_m$ from each level are fed back into the diffusion model to continue the denoising learning process.
Finally, after the progressive denoising learning process, a complete local map $Map_r$ is generated.

\subsection{Global Mapping: Elliptical Space Bayesian Fusion}
The local maps generated via diffusion refinement are capable of progressively recovering complete road structures from coarse initial maps. In consideration of the learning uncertainty inherent in neural networks, Bayesian optimization is adopted for local map fusion, with an elliptical distance propagated confidence module embedded to more effectively guide the fusion during the posterior update process.

The local map describes the road structure in a grid occupancy representation.  
First, the coordinate system of the local map is aligned to the global coordinate system according to the vehicle pose.
Compared with probabilistic mean-based fusion, the Bayesian approach possesses stronger evidence accumulation capability while circumventing numerical underflow issues in probability space.
Thus, the aligned local observations from individual frames are progressively accumulated into the global map via Bayesian recursive updating. 
Specifically, occupancy probability $p$ is represented as log-odds $L = \log(p / (1-p))$.
Then, the incremental update for the global map is formulated as:
\begin{equation}
L_t = L_{t-1} + \Delta L_t, \\
\end{equation}
\begin{equation}
\Delta L_t = \left( \log\frac{p^t_{\text{obs}}}{1-p^t_{\text{obs}}} - \log\frac{p_{\text{prior}}}{1-p_{\text{prior}}} \right),
\end{equation}
where $p^t_{\text{obs}}$ is the local observation probability and $p_{\text{prior}}$ is the static prior.
Due to significant disparities in the distribution density of different semantic categories, employing a uniform prior probability would systematically overestimate sparsely distributed categories. To address this, this section leverages empirical statistics from the observation sequence to independently set prior probabilities for different categories.

The Bayesian fusion framework is effective in accumulating multi-frame observations. However, it implicitly assumes a spatially uniform distribution of perceptual quality, overlooking the physical characteristic that, due to perspective projection and pixel resolution limitations of onboard cameras, prediction reliability degrades significantly at far-field regions. 
Inspired by human driving experience, observations in close proximity are generally more reliable, whereas those at farther distances exhibit greater uncertainty. This motivates the proposal of a distance-propagated confidence measure.

Considering that the perceptual range is asymmetrically distributed along the longitudinal and lateral directions, this section introduces an elliptical spatially normalized distance $d_n$ to provide a unified metric for the relative proximity of a local spatial point $(l_x,l_y)$. 
The rectangular perceptual region is projected into an elliptical space:
\begin{equation}
    d_n(l_x,l_y) = \sqrt{(\frac{l_x}{r_x})^2+(\frac{l_y}{r_y})^2},
\end{equation}
where $r_x$ and $y_x$ denote the lateral and longitudinal perceptual distances, respectively.
Based on the normalized distance, this section adopts the Cauchy function to transform it into a pixel-level adaptive weight $c(l_x,l_y)$. The Cauchy function possesses a smooth heavy-tailed characteristic, which effectively suppresses distant noise while preserving valid boundary information. It ensures a confidence of $1.0$ at the vehicle center and a natural decay to $0.5$ at the perceptual boundary:
\begin{equation}
    c(l_x,l_y) = \frac{1}{1+d_n^2}.
\end{equation}
Finally, the pixel-wise adaptive weights are applied to the incremental update learning process as follows: 
\begin{equation}
\Delta L_t^w(x) = \Delta L_t(x) \otimes c(l_x, l_y) \otimes f(P),
\end{equation}
where $c(l_x, l_y)$ is a confidence weight that decays with the elliptical distance from the ego-vehicle. $f(P)$ is a bilinear interpolation weight used to align local pixels to non-integer global grid positions. The final global map is recovered as:
\begin{equation}
\text{Map}_{\text{g}} = \sigma\left(L_0 + \sum_{t=1}^{T} \Delta L_t^w(x)\right).
\end{equation}
Through weighted Bayesian adaptive fusion, the limitations of conventional mean fusion in evidence accumulation and spatial perception quality modeling are effectively overcome, thereby providing a robust fusion framework for global map construction in large-scale dynamic scenarios.

\begin{table*}[t]
\fontsize{8}{7.8}\selectfont
\renewcommand{\arraystretch}{1.5}
\setlength\tabcolsep{9pt}
\caption{Quantitative results of BEV HD maps on nuScenes~\cite{caesar2020nuscenes} and Argoverse~\cite{wilson2023argoverse}. `\textbf{VT}' means view transformer, whereas `\textbf{OSM}' means OpenStreetMap. * represents data from the reference paper.}
\vspace{-1.5em}
\begin{center}
\resizebox{1.0\linewidth}{!}{\begin{tabular}{c|cc|cccc|cccc}
\hline
\multirow{2}{*}{\textbf{Method}} 
&\multirow{2}{*}{\textbf{VT}} 
&\multirow{2}{*}{\textbf{OSM}} 
&\multicolumn{4}{c|}{\textbf{nuScenes (60$\times$30)}}
&\multicolumn{4}{c}{\textbf{Argoverse (60$\times$30)}}\\
\cline{4-11}
& & & Div & Ped & Bound & mIoU(\%) & Div & Ped & Bound & mIoU(\%) \\ 
\hline
\hline
LSS*~\cite{philion2020lift} & Depth & & 39.5 &15.5 &40.7 &31.90 &31.4 &5.5 &26.2 &21.03\\
HDMapNet*~\cite{li2022hdmapnet} &MLP & &42.1 &21.1 & 42.8 &35.33 & 57.3 & 28.1 & 47.3 & 44.23 \\
DiffMap*~\cite{jia2024diffmap} &Diffusion & & 42.1 & 23.9 & 42.2 & 36.07 &-  &-  &-  &-  \\
P-MapNet~\cite{jiang2024p} &MLP & \CheckmarkBold & 44.3 & 23.3 & 43.8 & 37.13 & 53.5 & 30.1 & 47.3 & 43.63 \\
GenMapping~\cite{li2024genmapping} &IPM & \CheckmarkBold & - & - & - & - & 59.3 & 37.0 & 48.4 & 48.23 \\
 \textbf{Ours} & MLP/IPM & \CheckmarkBold &\textbf{45.9} & \textbf{25.8} & \textbf{45.4} & \textbf{39.03} & \textbf{60.4} & \textbf{37.0} & \textbf{49.4} & \textbf{48.93} \\
 \hline 
 \multirow{2}{*}{\textbf{Method}} 
&\multirow{2}{*}{\textbf{VT}} 
&\multirow{2}{*}{\textbf{OSM}} 
&\multicolumn{4}{c|}{\textbf{nuScenes (120$\times$60)}}
&\multicolumn{4}{c}{\textbf{nuScenes (240$\times$60)}}\\
\cline{4-11}
& & & Div & Ped & Bound & mIoU(\%) & Div & Ped & Bound & mIoU(\%) \\ 
\hline
\hline
 HDMapNet*~\cite{li2022hdmapnet} &MLP & & 39.20 & 23.00 & 39.10 & 33.77& 31.90 & 17.00 & 31.40 & 26.77 \\
 P-MapNet~\cite{jiang2024p} &MLP & \CheckmarkBold  & 45.50 & 30.90 & 46.20 & 40.87 & 49.00 & 40.90 & 46.60 & 45.50 \\
 Ours & MLP & \CheckmarkBold & \textbf{47.30} & \textbf{33.20} & \textbf{47.30} & \textbf{42.60} & \textbf{51.60} & \textbf{43.90} & \textbf{49.60} & \textbf{48.37} \\ \hline
\end{tabular}}
\end{center}
\label{tab:all}
\vspace{-1em}
\end{table*}
\begin{figure*}[t]
    \centering
    \includegraphics[width=0.83\textwidth]{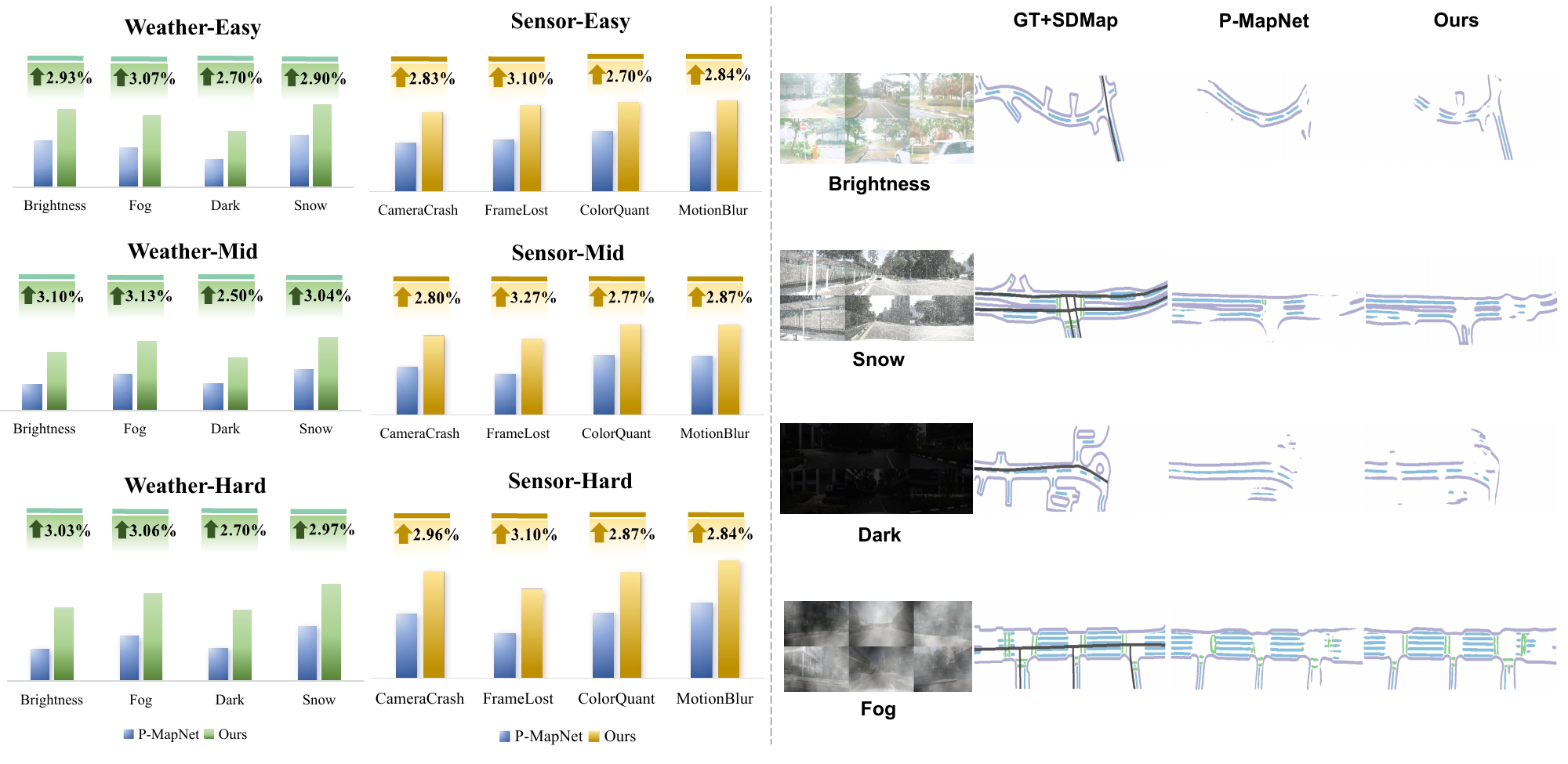}
    \caption{
    Performance of local mapping under extreme weather conditions with image corruptions. 
    }
    \label{fig:robust_merge}
    \vspace{-1.5em}
\end{figure*}
\subsection{The Overall Loss}
For the supervision of the proposed model, the focal loss~\cite{lin2017focal} is applied between the ground truth map $y_i$ and the predicted map $\hat{y}_i$:
\begin{equation}
\label{eq9}
Loss = \frac{1}{N} \sum_{i=1}^{N} \left( \text{BCE}(y_i, \hat{y}_i) \cdot  (1 - p_{t_i})^\gamma \right),
\end{equation}
where $\text{BCE}$ is the computation of binary cross-entropy loss, $p_{t_i}$ is the weighted sum of predicted probability and truth for each sample, and $N$ is the number of semantic classes.

\section{Experiments}

\subsection{Experimental Setup}
\noindent\textbf{Datasets.}
The nuScenes dataset~\cite{caesar2020nuscenes} is a large-scale autonomous driving benchmark comprising $1,000$ diverse scenes over a time span of $20s$. 
The Argoverse~\cite{wilson2023argoverse} dataset offers another rich autonomous driving repository, featuring high-definition maps and sensor data collected from six cities. 
The centerline map is sourced from OpenStreetMap~\cite{haklay2008openstreetmap}.

\noindent\textbf{Evaluation Metrics.}
The Intersection over Union (IoU) metric is applied as the validation criterion across all tasks.
Global IoU assesses global map quality by treating all map elements as a unified binary foreground. 
Higher values indicate more accurate skeleton recovery with increased robustness to local noise and fragmentation. It is defined as:
\begin{equation}
IoU_{Global} = \frac{TP_{all}}{TP_{all} + FP_{all} + FN_{all}}.
\end{equation}

\begin{table*}[t]
\centering 
\fontsize{8.5}{10}\selectfont 
\renewcommand{\arraystretch}{1.3} 
\setlength\tabcolsep{4pt} 
\caption{Comparison of global map construction methods across different cities.}
\label{tab:global_map_new}
\resizebox{1.0\linewidth}{!}{\begin{tabular}{l|l|cc|cc|cc|cc}
\hline
\multirow{2}{*}{\textbf{Global Fusion Method}} & \multirow{2}{*}{\textbf{Local Map Model}} & \multicolumn{2}{c|}{\textbf{singapore-onenorth}} & \multicolumn{2}{c|}{\textbf{singapore-hollandvillage}} & \multicolumn{2}{c|}{\textbf{singapore-queenstown}} & \multicolumn{2}{c}{\textbf{boston-seaport}} \\
\cline{3-10}
& & mIoU & Global IoU & mIoU & Global IoU & mIoU & Global IoU & mIoU & Global IoU  \\
\hline
\hline
\multirow{2}{*}{Direct Projection} & P-MapNet & 22.77 & 26.60 & 32.88 & 39.82 & 19.41 & 25.90 & 27.34 & 34.64  \\
& HPDR & 23.41 & 26.63 & 33.88 & 39.48 & 20.24 & 25.55 & 27.01 & 34.37  \\
\hline
\multirow{2}{*}{Probabilistic Mean} & P-MapNet & 16.60 & 22.48 & 30.18 & 41.27 & 12.99 & 20.06 & 24.00 & 30.91  \\
& HPDR & 17.92 & 23.70 & 32.19 & 42.91 & 14.03 & 21.52 & 25.24 & 32.09  \\
\hline
\multirow{2}{*}{\textbf{ESBF (Ours)}} & P-MapNet & 22.32 & 27.22 & 39.15 & 46.57 & 17.84 & 24.11 & 31.58 & 38.10  \\
& HPDR & \textbf{25.25} & \textbf{28.83} & \textbf{39.04} & \textbf{47.08} & \textbf{20.87} & \textbf{27.11} & \textbf{32.68} & \textbf{39.25} \\
\hline
\end{tabular}}
\vspace{-1em}
\end{table*}
\begin{figure*}[tb]
    \centering
    \includegraphics[width=0.75\textwidth]{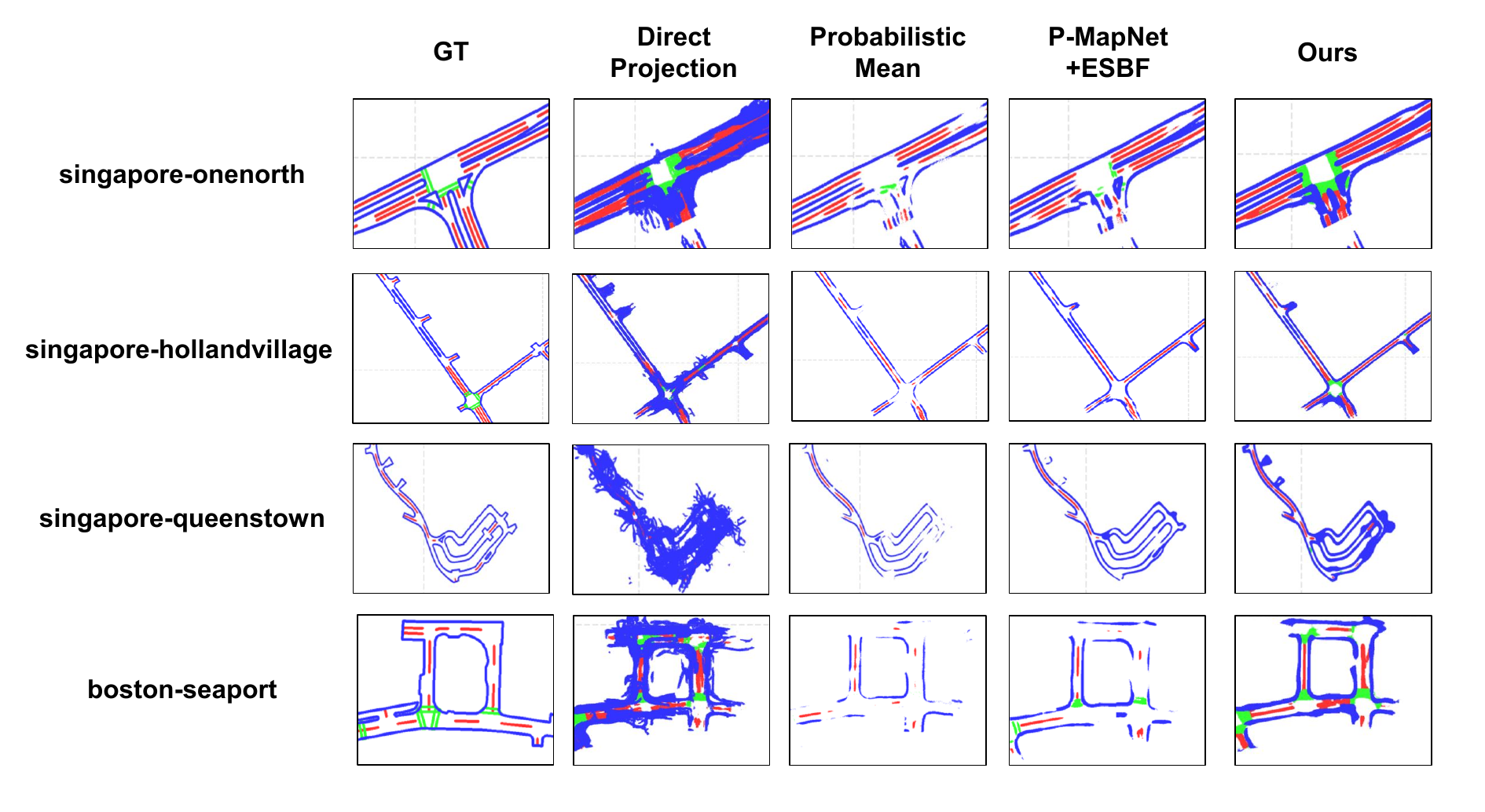}
    \caption{Local visualization details of global HD maps across four cities in the nuScenes dataset.}
    \label{fig:vis_global}
    \vskip -3ex
\end{figure*}
\noindent\textbf{Implementation Details.}
All experiments are trained on an RTX A6000 GPU using the Adam optimizer and StepLR scheduler. 
To mitigate systematic overestimation of sparse categories, the empirical prior probabilities are assigned: $0.1$ for \emph{Divider}, $0.02$ for \emph{Pedestrian}, and $0.09$ for \emph{Boundary}.
 
A two-stage training strategy is adopted for the HPDR module. 
First, the model is pre-trained for $20$ epochs using GT with $50\%$ random masking to simulate coarse priors. Subsequently, the model is fine-tuned for $10$ epochs using BEV features and coarse maps from the local mapping system. During inference, we initialize from Gaussian noise and employ the DDIM sampling rule to progressively denoise the latent distribution into a complete map representation across $t \in [1, 0]$.

\subsection{Comparison to State-of-the-art Methods}
\noindent \textbf{Local Mapping.}
Several competitive online map models are selected as baseline methods for comparison, including HDMapNet~\cite{li2022hdmapnet}, P-MapNet~\cite{jiang2024p}, and GenMapping~\cite{li2024genmapping}. 
The experimental results are presented in Table~\ref{tab:all}. For short-range local map construction, the proposed method achieves higher mapping accuracy on both the nuScenes and Argoverse datasets, improving by $ 1.90\%$ mIoU and $0.70\%$ mIoU, respectively.
Meanwhile, for longer-range local map construction, the proposed method achieves a $2.87\%$ mIoU improvement in accuracy. This benefit is attributed to the guidance of dual priors during the local refinement process, particularly the temporal prior knowledge, which provides effective guidance for more distant regions.

Cameras are highly susceptible to environmental disturbances. This section further evaluates the robustness of the HPDR module across distinct challenge scenarios.
As shown in Fig.~\ref{fig:robust_merge}, the proposed method outperforms P-MapNet~\cite{jiang2024p} across various complex scenarios.
Notably, under \emph{FrameLoss} conditions, leveraging historical prior information and road centerline guidance, our method realizes precision gains of $3.10\%$ mIoU, $3.27\%$ mIoU, and $3.10\%$ mIoU at \emph{easy}, \emph{medium}, and \emph{hard} difficulty levels, respectively.
Evidently, our method demonstrates stronger applicability in real-world environments.
Furthermore, our local refinement method demonstrates the capability to generate superior local maps from crowd-sourced data collected in diverse environments, thereby enabling accurate global map updates.

\noindent \textbf{Global Fusion.}
Several commonly adopted local map fusion methods for global map construction are selected as baselines for comparison in this section. One method directly accumulates instances after confidence filtering, while the other employs probabilistic additive fusion.
As shown in Table~\ref{tab:global_map_new}, our approach significantly outperforms these methods, yielding gains of $5.51\%$ Global IoU and $7.11\%$ mIoU over the probabilistic mean baseline. 
The global map is shown in Fig.~\ref{fig:vis_global}; our method suppresses noise and ghosting inherent in mean-based fusion, achieving sharper boundaries and consistent geometry where conventional baselines struggle.

\begin{table}[tb]
\fontsize{6}{6.5}\selectfont
\renewcommand{\arraystretch}{1.2}
\setlength\tabcolsep{9pt}
\centering
\caption{Ablation results on HPDR. }
\vskip -1ex
\label{tab:ablation_map_feat}
\resizebox{1.0\linewidth}{!}{\begin{tabular}{c|cc|c}
    \hline
    \textbf{Diffusion}&\textbf{Sparse features}& \textbf{Dense features} & \textbf{mIoU (\%)} \\
    \hline
    \hline
    \CheckmarkBold&\XSolidBrush &\XSolidBrush &41.33 \\
    \CheckmarkBold&\CheckmarkBold &\XSolidBrush &42.30 \\
    \CheckmarkBold&\CheckmarkBold &\CheckmarkBold &42.60\\
    \hline
\end{tabular}}
\vspace{-2em}
\end{table}
\begin{table}[tb]
\fontsize{6.5}{6.8}\selectfont
\renewcommand{\arraystretch}{1.2}
\setlength\tabcolsep{9pt}
\centering
\caption{Ablation results on ESBF. }
\vskip -1ex
\label{tab:ablation_module_wbaf}
\resizebox{1.0\linewidth}{!}{\begin{tabular}{l|cc}
    \hline
    \textbf{Method} & \textbf{Global IoU(\%)}  & \textbf{mIoU (\%)} \\
    \hline
    \hline
    Bayesian &35.06  &28.02 \\
    Bayesian+Bilinear &35.13  &28.14 \\
    Bayesian+Bilinear+Confidence &\textbf{35.57} &\textbf{29.46} \\
    \hline
\end{tabular}}
\vskip -3ex
\end{table}

\subsection{Ablation Studies}
\noindent\textbf{Ablation of HPDR Modules.} 
We evaluate the HPDR module on the $120\text{m}{\times}60\text{m}$ mapping task. 
As shown in Table~\ref{tab:ablation_map_feat}, performance improves incrementally with each component.
The prior guidance from temporal maps improves local refinement accuracy by $0.97\%$ mIoU, while the incorporation of dense BEV features further increases it by $0.30\%$ mIoU, enabling the recovery of finer details.
Notably, as centerline priors primarily inform drivable areas, their impact on crosswalk detection is negligible due to weak spatial correlation.

\noindent\textbf{Ablation of the ESBF module.}
This section evaluates the impact of bilinear interpolation and distance-based confidence weighting on the ESBF module. As shown in Table~\ref{tab:ablation_module_wbaf}, these modules effectively filter noisy distant observations while enhancing the reliability of near-field features.

\begin{table}[tb]
\fontsize{6}{6.8}\selectfont
\renewcommand{\arraystretch}{1.2}
\setlength\tabcolsep{9pt}
\centering
\caption{Ablation results of different prior learning methods on the HPDR module.}
\vskip-1ex
\label{tab:ablation_module_v2_1}
\resizebox{1.0\linewidth}{!}{\begin{tabular}{l|ccc|c}
    \hline
    \textbf{Method} & \textbf{Div.} & \textbf{Ped.} & \textbf{Bound.} & \textbf{mIoU (\%)} \\
    \hline
    \hline
    baseline &47.30 &32.30 &47.20 &42.27\\
    Concat &46.20 &31.60 &46.80 &41.53 \\
    Add &47.00 &32.70 &47.10 &42.27 \\
    Cross-Attn &\textbf{47.30} &\textbf{33.20} &\textbf{47.30} &\textbf{42.60}   \\
    \hline
\end{tabular}}
\vskip 2ex

\fontsize{6}{6.8}\selectfont
\renewcommand{\arraystretch}{1.2}
    \centering
    \caption{Ablation of diffusion sampling methods.}
    \vskip -1ex
    \label{tab:ablation_step}
    \resizebox{1.0\linewidth}{!}{\begin{tabular}{lc|ccc|c}
        \hline
        \textbf{Type} & \textbf{Step} & \textbf{Div.} & \textbf{Ped.} & \textbf{Bound.} & \textbf{mIoU (\%)} \\
        \hline
        \hline
        DDIM & 3 &47.30 &33.10 &47.40 &\textbf{42.60}  \\
        DDIM & 1 &47.30 &33.10 &47.30 &42.57   \\
        DDPM & 1000 &47.20 &33.20 &47.30 &42.57  \\
        \hline
    \end{tabular}}
\vskip-2ex
\end{table}
\begin{table}[!t]
\fontsize{7}{8}\selectfont
\renewcommand{\arraystretch}{1.2}
\setlength\tabcolsep{9pt}
\centering
\caption{Ablation results of different initial prior methods.}
\vskip -1ex
\label{tab:prior}
\resizebox{1.0\linewidth}{!}{\begin{tabular}{l|c|ccc}
\hline
\textbf{Global Fusion} &\textbf{Intial Prior Method} & \textbf{Global IoU} & \textbf{mIoU}    \\ \hline \hline
ESBF &Uniform method      & 27.90      & 20.50  \\
ESBF &Heuristic method     & \textbf{35.06}      & \textbf{28.02} \\ \hline 
\end{tabular}}
\vskip-2ex
\end{table}
\noindent\textbf{Ablation of the Prior Learning.}
During the HPDR module, different prior knowledge serves as conditional guidance for the diffusion model. This section discusses different embedding strategies.
As shown in Table~\ref{tab:ablation_module_v2_1}, the cross-attention mechanism achieves a state-of-the-art $42.60\%$ mIoU, surpassing feature addition and concatenation operations, respectively. 
Its advantage lies in the capacity for adaptive global learning, which effectively aligns the centerline with denoised features by capturing complex spatial relationships, thereby mitigating the inherent misalignment problem.

\noindent\textbf{Ablation of the Diffusion Sampling Methods.}
To evaluate the impact of diffusion sampling methods on local map refinement, ablation experiments are performed in Table~\ref{tab:ablation_step}. 
The sampling method from DDP~\cite{ji2023ddp} is employed to compare the efficiency of DDIM and DDPM. 
The results indicate that the model achieves the highest accuracy with only $3$ sampling steps. 
Even with a single step of generation, the model can achieve a sufficiently effective accuracy level, consistent with the expectations of DDP~\cite{ji2023ddp}.

\begin{table}[t]
\fontsize{9}{11}\selectfont
\renewcommand{\arraystretch}{1.2}
\setlength\tabcolsep{9pt}
\centering
\caption{The analysis results of map construction time.}
\label{tab:cost}
\resizebox{1.0\linewidth}{!}{\begin{tabular}{l|cccc}
\hline
\textbf{Local Methods}  &\textbf{Local mIoU(\%)} & \textbf{Global IoU(\%)} & \textbf{Time Costs}    \\ \hline \hline
P-MapNet~\cite{jiang2024p}      & 40.87      & 24.11 &15min  \\
Ours     & 42.60      & 27.11 &18min \\ \hline 
\end{tabular}}
\end{table}
\begin{table}[t]
\fontsize{6}{6.8}\selectfont
\renewcommand{\arraystretch}{1.2}
    \centering
    \caption{Results of closed-loop validation. 
    `ProbMap' denotes the global map fusion method based on probabilistic accumulation.
    `PMap' denotes the local map generation module based on the PMapNet model.}
    \label{tab:close}
    \resizebox{1.0\linewidth}{!}{\begin{tabular}{l|c|cccc}
        \hline
\textbf{ Task}   &\textbf{Method} & \textbf{Div} & \textbf{Ped} & \textbf{Bound} & \textbf{mIoU} \\   \hline  \hline
 \multirow{3}{*}{Sem} &   HDMapNet~\cite{li2022hdmapnet}             & 39.20 & 23.00 & 39.10 & 33.77 \\
   & HDMapNet+ProbMap     & 52.40 & 37.80 & 55.20 & 48.47 \\
   
    &HDMapNet+L2G-Map       & \textbf{54.80} & \textbf{44.10} & \textbf{57.20} & \textbf{52.03} \\
    \hline 
    \textbf{Task}   &\textbf{Method} & \textbf{Div} & \textbf{Ped} & \textbf{Bound} & \textbf{mAP} \\   \hline  \hline
\multirow{3}{*}{\textbf{Vec}}  &StreamMapNet~\cite{yuan2024streammapnet}   &61.70       & 66.30      & 62.10      & 63.40 \\
&StreamMapNet+PMap &89.13 &96.25 &58.97 &81.45 \\
&StreamMapNet+L2G-Map &\textbf{90.56} &\textbf{95.04} &\textbf{64.71} &\textbf{83.44} \\ \hline
\end{tabular}}
\vskip -1em
\end{table}
\noindent\textbf{Ablation of Initial Probability in ESBF.}
Given the imbalanced distribution between map classes, this section compares different probability initialization schemes. 
Scheme I adopts a uniform prior constraint of $50\%$. 
Scheme II employs the proposed heuristic method. 
The ablation results presented in Table~\ref{tab:prior} demonstrate that the data-driven heuristic prior better accommodates the sparse characteristics of real-world scenarios, thereby yielding more precise update increments.
\begin{figure}[!t]
    \centering
    \includegraphics[scale=0.47]{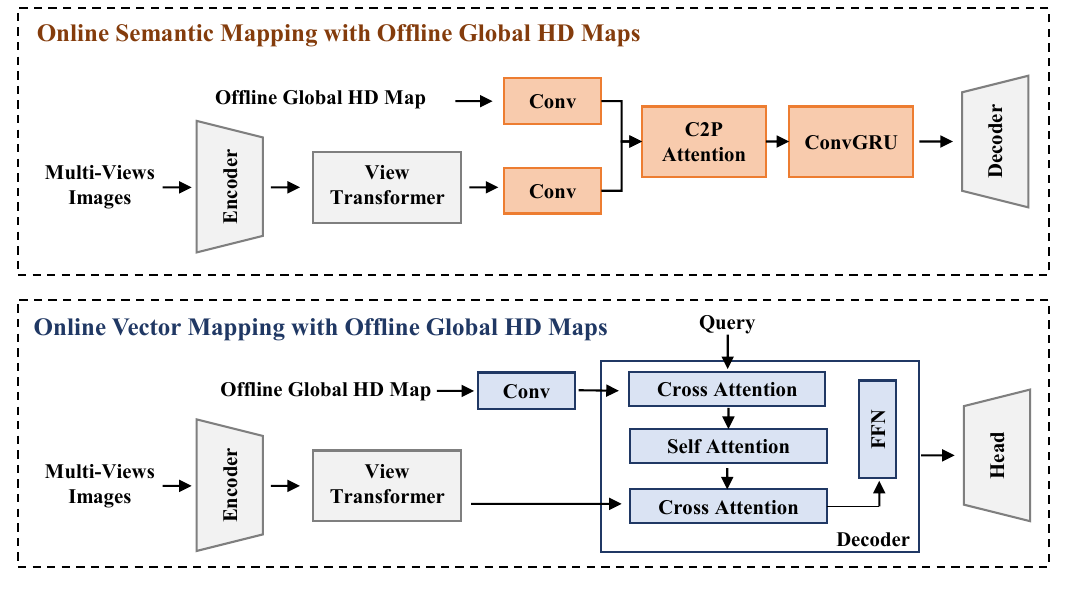}
    \vspace{-0.6em}
    \caption{The framework of closed-loop verification for online semantic mapping and vector mapping tasks.}
    \label{fig:close}
    \vspace{-2em}
\end{figure}

\subsection{Efficiency Analysis of Global Map Construction}
To evaluate the computational efficiency of global map construction, we conduct additional experiments on the nuScenes validation set, which contains $6019$ frames. 
In this evaluation, only the local mapping module is replaced, with ESBF uniformly employed for global fusion.
As shown in Table~\ref{tab:cost}, while the baseline method takes approximately $15$ minutes, our approach requires about $18$ minutes.
As the proposed L2G-Map solution targets offline HD map construction and yields substantial accuracy improvements, we consider this computational overhead to be acceptable.

\subsection{Closed-Loop Verification of the Global Map}
The offline global HD map, serving as a powerful prior, can be embedded into online mapping models. 
Thus, closed-loop tests are conducted for online semantic mapping and vector mapping tasks.
Specifically, HDMapNet~\cite{li2022hdmapnet} is adopted as the baseline for semantic mapping, while StreamMapNet~\cite{yuan2024streammapnet} serves as the baseline for vector mapping.

As shown in Fig.~\ref{fig:close}, the offline global HD map is incorporated into each baseline framework. Specifically, in online semantic mapping, deep features derived from the global map are fused into BEV representations. In vector mapping, the global HD map is leveraged as a prior to guide instance-level decoding.
The experimental results are presented in Table~\ref{tab:close}. 
Evidently, the prior guidance provided by the global map leads to a substantial improvement in online map perception accuracy.
Specifically, the semantic mapping accuracy is enhanced by $18.26\%$ mIoU, while the vector mapping task achieves a $20.00\%$ mAP improvement.
Furthermore, in comparison with other methods, the proposed approach demonstrates more pronounced advantages.

\subsection{Verification of HPDR in Real-World Environments}
\begin{figure}[!t]
    \centering
    \includegraphics[scale=0.6]{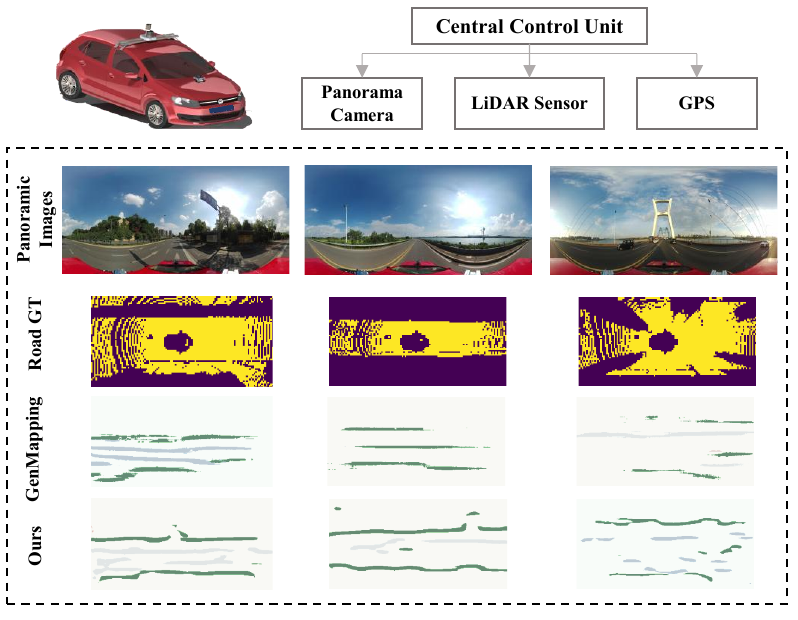}
    \vspace{-0.6em}
    \caption{Verification results on real-world sequences. The panoramic image data were collected on urban main roads.}
    \label{fig:real}
    \vspace{-1.5em}
\end{figure}

To further verify the applicability of the proposed method, this section conducts experiments in real-world environments.
Real-world sequence data are collected on urban roads using a vehicle equipped with a panoramic camera, a LiDAR, and GPS sensors for comprehensive 360{\textdegree} surrounding scene parsing.
Since this work focuses on vision-based mapping using multi-view inputs, we employ only the panoramic images for local map generation in this validation, while LiDAR and GPS data are primarily used for ground-truth reference and pose alignment.
HPDR is first pretrained on Argoverse, with GenMapping~\cite{li2024genmapping} adopted as the local mapping model.
Subsequently, the panoramas are directly fed into the pretrained model in place of multi-view images to generate refined local maps.
As shown in Fig.~\ref{fig:real}, despite the favorable generalization capability of GenMapping, it fails to fully recognize the complete road structures in local maps.
Upon refinement by our proposed HPDR, the local maps exhibit substantial quality improvement, recovering full road structures with enhanced completeness and fidelity.
The incomplete segmentation results, attributable to the geometric discrepancies between panoramic and pinhole images, highlight the need for further research on cross-domain generalization.
Furthermore, for panoramic images with a resolution of $256{\times}1240$, the refinement module processes each frame in about $0.13s$, which is sufficiently efficient for offline mapping in large-scale urban environments.

\section{Conclusion}

In this paper, we have presented L2G-Map, a novel local-to-global mapping framework that addresses the challenges of incomplete local map generation and suboptimal global map fusion in complex driving environments. The designed hierarchical prior diffusion refinement module leverages multi-level priors to guide the diffusion process, progressively recovering coherent local maps from coarse initial predictions. The proposed elliptical space Bayesian fusion module incorporates a confidence estimation strategy based on elliptical distance propagation, achieving probabilistically optimal aggregation of multi-frame observations.
Extensive experiments conducted on public datasets, as well as real-world urban panoramic data, demonstrate that the proposed method consistently outperforms competitive baselines in both local map refinement and global map construction.

For future work, we plan to investigate map models specifically tailored to panoramic imagery and extend the proposed framework to more challenging scenarios such as adverse weather conditions and unstructured road environments.

\bibliographystyle{IEEEtran}
\bibliography{bib.bib}

\begin{thebibliography}{10}
\providecommand{\url}[1]{#1}
\csname url@samestyle\endcsname
\providecommand{\newblock}{\relax}
\providecommand{\bibinfo}[2]{#2}
\providecommand{\BIBentrySTDinterwordspacing}{\spaceskip=0pt\relax}
\providecommand{\BIBentryALTinterwordstretchfactor}{4}
\providecommand{\BIBentryALTinterwordspacing}{\spaceskip=\fontdimen2\font plus
\BIBentryALTinterwordstretchfactor\fontdimen3\font minus \fontdimen4\font\relax}
\providecommand{\BIBforeignlanguage}[2]{{%
\expandafter\ifx\csname l@#1\endcsname\relax
\typeout{** WARNING: IEEEtran.bst: No hyphenation pattern has been}%
\typeout{** loaded for the language `#1'. Using the pattern for}%
\typeout{** the default language instead.}%
\else
\language=\csname l@#1\endcsname
\fi
#2}}
\providecommand{\BIBdecl}{\relax}
\BIBdecl

\bibitem{mappath}
Z.~Gao \emph{et~al.}, ``Learning implicit map representations from trajectories: An enhanced map-free framework for motion forecasting,'' in \emph{SMC}, 2025.

\bibitem{yuan2026unimapgen}
Y.~Yuan \emph{et~al.}, ``{UniMapGen:} {A} generative framework for large-scale map construction from multi-modal data,'' in \emph{AAAI}, 2026.

\bibitem{philion2020lift}
J.~Philion and S.~Fidler, ``Lift, splat, shoot: Encoding images from arbitrary camera rigs by implicitly unprojecting to {3D},'' in \emph{ECCV}, 2020.

\bibitem{li2022hdmapnet}
Q.~Li \emph{et~al.}, ``{HDMapNet:} {An} online {HD} map construction and evaluation framework,'' in \emph{ICRA}, 2022.

\bibitem{jiang2024p}
Z.~Jiang \emph{et~al.}, ``{P-MapNet:} {Far-Seeing} map generator enhanced by both {SDMap} and {HDMap} priors,'' \emph{IEEE Robotics and Automation Letters}, 2024.

\bibitem{li2024genmapping}
S.~Li \emph{et~al.}, ``{GenMapping:} {Unveiling} the potential of inverse perspective mapping for robust online {HD} map construction,'' \emph{IEEE Open Journal of Intelligent Transportation Systems}, 2026.

\bibitem{li2024bevformer}
Z.~Li \emph{et~al.}, ``{BEVFormer:} {Learning} bird's-eye-view representation from {LiDAR-camera} via spatiotemporal transformers,'' \emph{IEEE Transactions on Pattern Analysis and Machine Intelligence}, 2025.

\bibitem{haklay2008openstreetmap}
M.~Haklay and P.~Weber, ``{OpenStreetMap:} {User-generated} street maps,'' \emph{IEEE Pervasive Computing}, 2008.

\bibitem{wu2024blos}
H.~Wu \emph{et~al.}, ``{BLOS-BEV:} {Navigation} map enhanced lane segmentation network, beyond line of sight,'' in \emph{IV}, 2024.

\bibitem{xiong2023neural}
X.~Xiong \emph{et~al.}, ``Neural map prior for autonomous driving,'' in \emph{CVPR}, 2023.

\bibitem{shi2024globalmapnet}
A.~Shi \emph{et~al.}, ``{GlobalMapNet:} {An} online framework for vectorized global {HD} map construction,'' \emph{arXiv preprint arXiv:2409.10063}, 2024.

\bibitem{yuan2024streammapnet}
T.~Yuan \emph{et~al.}, ``{StreamMapNet:} {Streaming} mapping network for vectorized online {HD} map construction,'' in \emph{WACV}, 2024.

\bibitem{li2024dtclmapper}
S.~Li \emph{et~al.}, ``{DTCLMapper:} {Dual} temporal consistent learning for vectorized {HD} map construction,'' \emph{IEEE Transactions on Intelligent Transportation Systems}, 2024.

\bibitem{li2026amap}
R.~Li \emph{et~al.}, ``{AMap:} {Distilling} future priors for ahead-aware online {HD} map construction,'' in \emph{CVPR}, 2026.

\bibitem{kingma2013auto}
D.~P. Kingma and M.~Welling, ``Auto-encoding variational bayes,'' in \emph{ICLR}, 2014.

\bibitem{vahdat2020nvae}
A.~Vahdat and J.~Kautz, ``{NVAE:} {A} deep hierarchical variational autoencoder,'' in \emph{NeurIPS}, 2020.

\bibitem{he2022masked}
K.~He \emph{et~al.}, ``Masked autoencoders are scalable vision learners,'' in \emph{CVPR}, 2022.

\bibitem{liu2205mixed}
J.~Liu \emph{et~al.}, ``{MixMIM:} {Mixed} and masked image modeling for efficient visual representation learning,'' \emph{arXiv preprint arXiv:2205.13137}, 2022.

\bibitem{zhu2023mapprior}
X.~Zhu \emph{et~al.}, ``{MapPrior:} {Bird's-eye view} map layout estimation with generative models,'' in \emph{ICCV}, 2023.

\bibitem{dhariwal2021diffusion}
P.~Dhariwal and A.~Nichol, ``Diffusion models beat {GANs} on image synthesis,'' in \emph{NeurIPS}, 2021.

\bibitem{ho2020denoising}
J.~Ho \emph{et~al.}, ``Denoising diffusion probabilistic models,'' in \emph{NeurIPS}, 2020.

\bibitem{rombach2022high}
R.~Rombach \emph{et~al.}, ``High-resolution image synthesis with latent diffusion models,'' in \emph{CVPR}, 2022.

\bibitem{zhang2023adding}
L.~Zhang \emph{et~al.}, ``Adding conditional control to text-to-image diffusion models,'' in \emph{ICCV}, 2023.

\bibitem{ji2023ddp}
Y.~Ji \emph{et~al.}, ``{DDP:} {Diffusion} model for dense visual prediction,'' in \emph{ICCV}, 2023.

\bibitem{jia2024diffmap}
P.~Jia \emph{et~al.}, ``{DiffMap:} {Enhancing} map segmentation with map prior using diffusion model,'' \emph{IEEE Robotics and Automation Letters}, 2024.

\bibitem{ross2022bev}
J.~Ross \emph{et~al.}, ``{BEV-SLAM:} {Building} a globally-consistent world map using monocular vision,'' in \emph{IROS}, 2022.

\bibitem{qiao2025csmapping}
Z.~Qiao \emph{et~al.}, ``{CSMapping:} {Scalable} crowdsourced semantic mapping and topology inference for autonomous driving,'' \emph{arXiv preprint arXiv:2512.03510}, 2025.

\bibitem{jaderberg2015spatial}
M.~Jaderberg \emph{et~al.}, ``Spatial transformer networks,'' in \emph{NeurIPS}, 2015.

\bibitem{caesar2020nuscenes}
H.~Caesar \emph{et~al.}, ``{nuScenes:} {A} multimodal dataset for autonomous driving,'' in \emph{CVPR}, 2020.

\bibitem{wilson2023argoverse}
B.~Wilson \emph{et~al.}, ``{Argoverse 2:} {Next} generation datasets for self-driving perception and forecasting,'' in \emph{NeurIPS}, 2021.

\bibitem{lin2017focal}
T.-Y. Lin \emph{et~al.}, ``Focal loss for dense object detection,'' in \emph{ICCV}, 2017.

\end{thebibliography}

\end{document}